\documentclass{article} 
\usepackage{iclr2017_conference,times}

\usepackage[utf8]{inputenc} 	
\usepackage[T1]{fontenc}    	
\usepackage{hyperref}       	
\usepackage{url}            	
\usepackage{booktabs}       	
\usepackage{amsfonts}       	
\usepackage{nicefrac}       	
\usepackage{microtype}      	
\usepackage{bm} 				
\usepackage{comment}			
\usepackage{algorithm}			
\usepackage{algpseudocode}
\usepackage{graphicx} 			
\usepackage{color} 				
\usepackage{amsmath, amsfonts} 	

\newcommand{\xbacki}{\mathbf{x}_{\backslash i}}	

\title{Visualizing Deep Neural Network Decisions: \\ Prediction Difference Analysis}

\author{Luisa M Zintgraf \\
University of Amsterdam \\
\texttt{lmzintgraf@gmail.com} \\
\And
\hspace{-7.55cm} Taco S Cohen \\
\hspace{-7.55cm} University of Amsterdam \\
\hspace{-7.55cm} \texttt{t.s.cohen@uva.nl}
\AND
Tameem Adel \\
University of Amsterdam \\
\texttt{tameem.hesham@gmail.com} \\
\And
Max Welling \\
University of Amsterdam \\
Canadian Institute of Advanced Research \\
\texttt{m.welling@uva.nl}
}

\author{Luisa M Zintgraf$^{1,3}$, Taco S Cohen$^1$, Tameem Adel$^1$, Max Welling$^{1,2}$\\
$^1$University of Amsterdam, 
$^2$Canadian Institute of Advanced Research, 
$^3$Vrije Universiteit Brussel \\
\texttt{\{lmzintgraf,tameem.hesham\}@gmail.com}, \texttt{\{t.s.cohen, m.welling\}@uva.nl}
}

\iclrfinalcopy

\begin{document}

\maketitle

\begin{abstract}
This article presents the \textit{prediction difference analysis} method for visualizing the response of a deep neural network to a specific input. When classifying images, the method highlights areas in a given input image that provide evidence for or against a certain class. It overcomes several shortcoming of previous methods and provides great additional insight into the decision making process of classifiers. Making neural network decisions interpretable through visualization is important both to improve models and to accelerate the adoption of black-box classifiers in application areas such as medicine. We illustrate the method in experiments on natural images (ImageNet data), as well as medical images (MRI brain scans).
\end{abstract}


\section{Introduction} \label{introduction}

\begin{figure*}[b]
\includegraphics[width=0.4\textwidth]{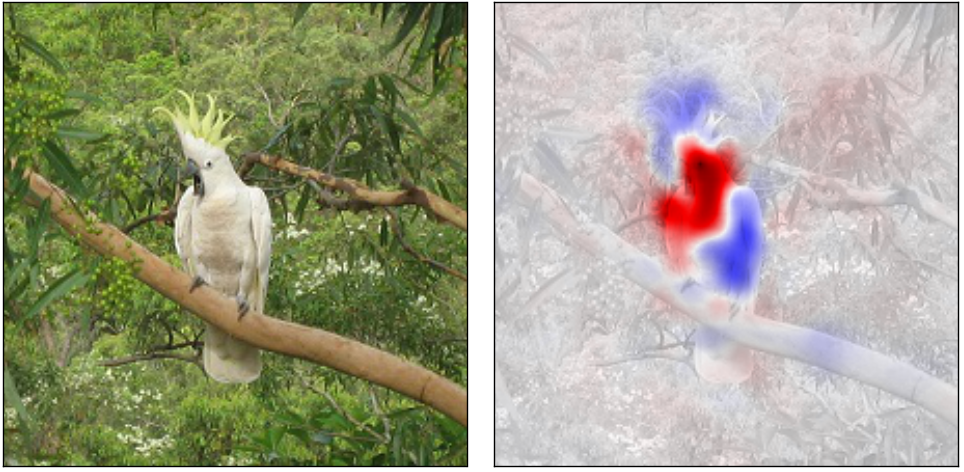}
\centering
\caption{\small Example of our visualization method: explains why the DCNN (GoogLeNet) predicts "cockatoo". Shown is the evidence for (red) and against (blue) the prediction. We see that the facial features of the cockatoo are most supportive for the decision, and parts of the body seem to constitute evidence against it. In fact, the classifier most likely considers them evidence for the second-highest scoring class, white wolf.}
\label{fig:intro_example}
\end{figure*}

Over the last few years, deep neural networks (DNNs) have emerged as the method of choice for perceptual tasks such as speech recognition and image classification. In essence, a DNN is a highly complex non-linear function, which makes it hard to understand how a particular classification comes about. This lack of transparency is a significant impediment to the adoption of deep learning in areas of industry, government and healthcare where the cost of errors is high.

In order to realize the societal promise of deep learning - e.g., through self-driving cars or personalized medicine - it is imperative that classifiers learn to explain their decisions, whether it is in the lab, the clinic, or the courtroom. In scientific applications, a better understanding of the complex dependencies learned by deep networks could lead to new insights and theories in poorly understood domains.

In this paper, we present a new, probabilistically sound methodology for explaining classification decisions made by deep neural networks. The method can be used to produce a saliency map for each (instance, node) pair that highlights the parts (features) of the input that constitute most evidence for or against the activation of the given (internal or output) node. See figure \ref{fig:intro_example} for an example.

In the following two sections, we review related work and then present our approach. In section \ref{sec:experiments} we provide several demonstrations of our technique for deep convolutional neural networks (DCNNs) trained on ImageNet data, and further how the method can be applied when classifying MRI brain scans of HIV patients with neurodegenerative disease.


\section{Related Work}\label{sec:related}

Broadly speaking, there are two approaches for understanding DCNNs through visualization investigated in the literature:
find an input image that \textit{maximally activates} a given unit or class score to visualize what the network is looking for \citep{erhan2009visualizing, simonyan2013deep, yosinski2015understanding}, or
visualize how the network responds to a \textit{specific} input image in order to explain a particular classification made by the network. The latter will be the subject of this paper.

One such instance-specific method is \textit{class saliency visualization} proposed by \citet{simonyan2013deep} who measure how sensitive the classification score is to small changes in pixel values, by computing the partial derivative of the class score with respect to the input features using standard backpropagation. They also show that there is a close connection to using deconvolutional networks for visualization, proposed by \citet{zeiler2014visualizing}.
Other methods include \citet{shrikumar2016not}, who compare the activation of a unit when a specific input is fed forward through the net to a reference activation for that unit. \citet{zhou2016learning} and \citet{bach2015pixel} also generate interesting visualization results for individual inputs, but are both not as closely related to our method as the two papers mentioned above.
The idea of our method is similar to another analysis \citet{zeiler2014visualizing} make: they estimate the importance of input pixels by visualizing the probability of the (correct) class as a function of a gray patch occluding parts of the image. In this paper, we take a more rigorous approach at both removing information from the image and evaluating the effect of this.

In the field of medical image classification specifically, a widely used method for visualizing feature importances is to simply plot the weights of a linear classifier \citep{kloppel2008automatic, ecker2010describing}, or the p-values of these weights (determined by permutation testing) \citep{mourao2005classifying, wang2007support}.
These are independent of the input image, and, as argued by \citet{gaonkar2013analytic} and \citet{haufe2014interpretation}, interpreting these weights can be misleading in general.

The work presented in this paper is based on an instance-specific method by \citet{robnik2008explaining}, the \textit{prediction difference analysis}, which is reviewed in the next section. 
Our main contributions are three substantial improvements of this method: \emph{conditional sampling} (section \ref{sec:approach:conditional}), \emph{multivariate analysis} (section \ref{sec:approach:multivariate_analysis}), and \emph{deep visualization} (section \ref{sec:approach:deepvis}).


\section{Approach} \label{sec:approach}

Our method is based on the technique presented by \cite{robnik2008explaining}, which we will now review. For a given prediction, the method assigns a \textit{relevance} value to each input feature with respect to a class $c$. The basic idea is that the relevance of a feature $x_i$ can be estimated by measuring how the prediction changes if the feature is unknown, i.e., the difference between $p(c|\mathbf{x})$ and $p(c|\xbacki)$, where $\xbacki$ denotes the set of all input features except $x_i$.

To find $p(c | \xbacki )$, i.e., evaluate the prediction when a feature is unknown, the authors propose three strategies.
The first is to label the feature as unknown (which only few classifiers allow).
The second is to re-train the classifier with the feature left out (which is clearly infeasible for DNNs and high-dimensional data like images).
The third approach is to \textit{simulate} the absence of a feature by marginalizing the feature:
\begin{equation} \label{eq:marginal-exact}
p(c|\xbacki) = \sum\limits_{x_i} p(x_i|\xbacki) p(c|\xbacki, x_i ) \ 
\end{equation}
(with the sum running over all possible values for $x_i$). However, modeling $p(x_i|\xbacki)$ can easily become infeasible with a large number of features. Therefore, the authors approximate equation (\ref{eq:marginal-exact}) by assuming that feature $x_i$ is independent of the other features, $\xbacki$:
\begin{equation} \label{eq:marginal-approx}
p(c|\xbacki) \approx \sum\limits_{x_i} p(x_i) p(c|\xbacki, x_i) \ .
\end{equation}
The prior probability $p(x_i)$ is usually approximated by the empirical distribution for that feature. 

Once the class probability $p(c|\xbacki)$ is estimated, it can be compared to $p(c|\mathbf{x})$. We stick to an evaluation proposed by the authors referred to as \textit{weight of evidence}, given by
\begin{equation} \label{eq:prediction-difference}
\text{WE}_i(c|\mathbf{x})=\log_2\left(\text{odds}(c|\mathbf{x})\right)-\log_2\left(\text{odds}(c|\xbacki)\right) ,
\end{equation}
where $\text{odds}(c|\mathbf{x}) = p(c|\mathbf{x}) / (1-p(c|\mathbf{x}))$.
To avoid problems with zero probabilities, Laplace correction $\displaystyle p\leftarrow (pN+1)/(N+K)$ is used, where $N$ is the number of training instances and $K$ the number of classes.

The method produces a relevance vector $(\text{WE}_i)_{i=1 \ldots m}$ ($m$ being the number of features) of the same size as the input, which reflects the relative importance of all features. A large prediction difference means that the feature contributed substantially to the classification, whereas a small difference indicates that the feature was not important for the decision.
A positive value $\text{WE}_i$ means that the feature has contributed evidence \textit{for} the class of interest: removing it would decrease the confidence of the classifier in the given class. A negative value on the other hand indicates that the feature displays evidence \textit{against} the class: removing it also removes potentially conflicting or irritating information and the classifier becomes more certain in the investigated class.

\begin{figure}
\includegraphics[width=0.4\textwidth]{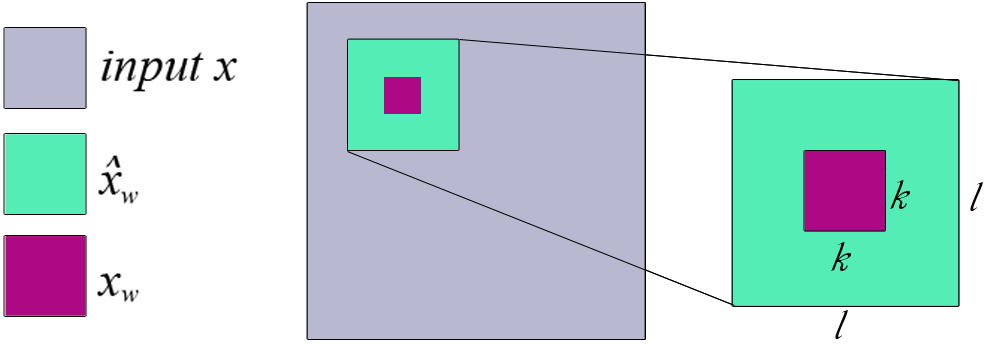}
\centering
\caption{\small Simple illustration of the sampling procedure in algorithm \ref{alg:pred-diff}. Given the input image $x$, we select every possible patch $x_w$ (in a sliding window fashion) of size $k\times k$ and place a larger patch $\hat{x}_w$ of size $l\times l$ around it. We can then conditionally sample $x_w$ by conditioning on the surrounding patch $\hat{x}_w$.}
\label{fig:box-visualisation}
\end{figure}

\begin{algorithm}[bt]
   \caption{Evaluating the prediction difference using conditional and multivariate sampling}
   \label{alg:pred-diff}
\begin{algorithmic}
   \State \textbf{Input:} classifier with outputs p(c|x), input image $\mathbf{x}$ of size $n \times n$, inner patch size $k$, outer patch size $l>k$, class of interest $c$, probabilistic model over patches of size $l \times l$, number of samples $S$
   \State \textbf{Initialization:} WE = zeros(n*n), counts = zeros(n*n)
   \For{ every patch $\mathbf{x}_w$ of size $k\times k$ {\bfseries in} $\mathbf{x}$}
     \State $\mathbf{x}' = \text{copy}(\mathbf{x})$
     \State $\text{sum}_w = 0$
   	 \State define patch $\hat{\mathbf{x}}_{w}$ of size $l\times l$ that contains $\mathbf{x}_w$
     \For{$s=1$ {\bfseries to} $S$}
       \State $\mathbf{x}'_w \leftarrow\mathbf{x}_w$ sampled from $p(\mathbf{x}_w|\hat{\mathbf{x}}_{w}\backslash \mathbf{x}_w)$
       \State $\text{sum}_w \mathrel{+}= p(c|\mathbf{x}')$ \Comment evaluate classifier
     \EndFor
     \State $p(c|\mathbf{x}\backslash \mathbf{x}_w) := \text{sum}_w / S$
   	 \State WE[coordinates of $\mathbf{x}_w$] $\mathrel{+}= \log_2(\text{odds}(c|\mathbf{x})) - \log_2(\text{odds}(c|\mathbf{x}\backslash \mathbf{x}_w))$
     \State counts[coordinates of $\mathbf{x}_w$] $\mathrel{+}= 1$
   \EndFor
   \State \textbf{Output:} WE / counts \Comment point-wise division
\end{algorithmic}
\end{algorithm}

\subsection{Conditional Sampling} \label{sec:approach:conditional}

In equation (\ref{eq:prediction-difference}), the conditional probability $p(x_i|\xbacki)$ of a feature $x_i$ is approximated using the marginal distribution $p(x_i)$. This is a very crude approximation. In images for example, a pixel's value is highly dependent on other pixels.
We propose a much more accurate approximation, based on the following two observations: a pixel depends most strongly on a small neighborhood around it, and the conditional of a pixel given its neighborhood does not depend on the position of the pixel in the image.
For a pixel $x_i$, we can therefore find a patch $\hat{\mathbf{x}}_i$ of size $l \times l$ that contains $x_i$, and condition on the remaining pixels in that patch:
\begin{equation} \label{eq:conditional-approx}
p(x_i|\xbacki) \approx p(x_i|\hat{\mathbf{x}}_{\backslash i}) \ .
\end{equation}
This greatly improves the approximation while remaining completely tractable.

For a feature to become relevant when using conditional sampling, it now has to satisfy two conditions: being relevant to predict the class of interest, and be hard to predict from the neighboring pixels. Relative to the marginal method, we therefore downweight the pixels that can easily be predicted and are thus redundant in this sense.

\subsection{Multivariate Analysis} \label{sec:approach:multivariate_analysis}

\citet{robnik2008explaining} take a \textit{univariate} approach: only one feature at a time is removed. However, we would expect that a neural network is relatively robust to just one feature of a high-dimensional input being unknown, like a pixel in an image. Therefore, we will remove several features at once by again making use of our knowledge about images by strategically choosing these feature sets: patches of connected pixels. Instead of going through all individual pixels, we go through all patches of size $k\times k$ in the image ($k\times k\times 3$ for RGB images and $k\times k\times k$ for 3D images like MRI scans), implemented in a sliding window fashion. The patches are overlapping, so that ultimately an individual pixel's relevance is obtained by taking the average relevance obtained from the different patches it was in.

Algorithm \ref{alg:pred-diff} and figure \ref{fig:box-visualisation} illustrate how the method can be implemented, incorporating the proposed improvements.

\subsection{Deep Visualization of Hidden Layers} \label{sec:approach:deepvis}

When trying to understand neural networks and how they make decisions, it is not only interesting to analyze the input-output relation of the classifier, but also to look at what is going on inside the hidden layers of the network. We can adapt the method to see how the units of any layer of the network influence a node from a deeper layer. 
Mathematically, we can formulate this as follows. Let $\mathbf{h}$ be the vector representation of the values in a layer $H$ in the network (after forward-propagating the input up to this layer).
Further, let $z = z(\mathbf{h})$ be the value of a node that depends on $\mathbf{h}$, i.e., a node in a subsequent layer.
Then the analog of equation (\ref{eq:marginal-approx}) is given by the expectation:
\begin{equation}
	g(z|\mathbf{h}_{\backslash i})
	\equiv \mathbb{E}_{p(h_i|\mathbf{h}_{\backslash i})} \left[ z(\mathbf{h}) \right] 
	= \sum\limits_{h_i} p(h_i|\mathbf{h}_{\backslash i} ) z( \mathbf{h}_{\backslash i}, h_i ) \ ,
\end{equation}
which expresses the distribution of $z$ when unit $h_i$ in layer $H$ is unobserved.
The equation now works for arbitrary layer/unit combinations, and evaluates to the same as equation (\ref{eq:marginal-exact}) when the input-output relation is analyzed.
To evaluate the difference between $g(z|\mathbf{h})$ and $g(z|\mathbf{h}_{\backslash i})$, we will in general use the \textit{activation difference}, 
$ \text{AD}_i(z|\mathbf{h}) = g(z|\mathbf{h}) - g(z|\mathbf{h}_{\backslash i}) \ , $
for the case when we are not dealing with probabilities (and equation (\ref{eq:prediction-difference}) is not applicable).


\section{Experiments} \label{sec:experiments}

In this section, we illustrate how the proposed visualization method can be applied, on the ImageNet dataset of natural images when using DCNNs (section \ref{sec:experiments:imagenet}), and on a medical imaging dataset of MRI scans when using a logistic regression classifier (section \ref{sec:experiments:mri}). 
For marginal sampling we always use the empirical distribution, i.e., we replace a feature (patch) with samples taken directly from other images, at the same location. For conditional sampling we use a multivariate normal distribution. 
For both sampling methods we use $10$ samples to estimate $p(c|\xbacki)$ (since no significant difference was observed with more samples). Note that all images are best viewed digital and in color.

Our implementation is available at \url{github.com/lmzintgraf/DeepVis-PredDiff}.

\subsection{ImageNet: Understanding how a DCNN makes decisions} \label{sec:experiments:imagenet}

We use images from the ILSVRC challenge \citep{ILSVRC15} (a large dataset of natural images from  $1000$ categories) and three DCNNs: the AlexNet \citep{krizhevsky2012imagenet}, the GoogLeNet \citep{szegedy2015going} and the (16-layer) VGG network \citep{simonyan2014very}. We used the publicly available pre-trained models that were implemented using the deep learning framework caffe \citep{jia2014caffe}. 
Analyzing one image took us on average 20, 30 and 70 minutes for the respective classifiers AlexNet, GoogLeNet and VGG (using the GPU implementation of caffe and mini-batches with the standard settings of $10$ samples and a window size of $k=10$). 

The results shown here are chosen from among a small set of images in order to show a range of behavior of the algorithm. The shown images are quite representative of the performance of the method in general. Examples on randomly selected images, including a comparison to the sensitivity analysis of \cite{simonyan2013deep}, can be seen in appendix \ref{app:random_results}.

We start this section by demonstrating our proposed improvements (sections \ref{sec:approach:conditional} - \ref{sec:approach:deepvis}).

\begin{figure}
\includegraphics[width=0.95\textwidth]{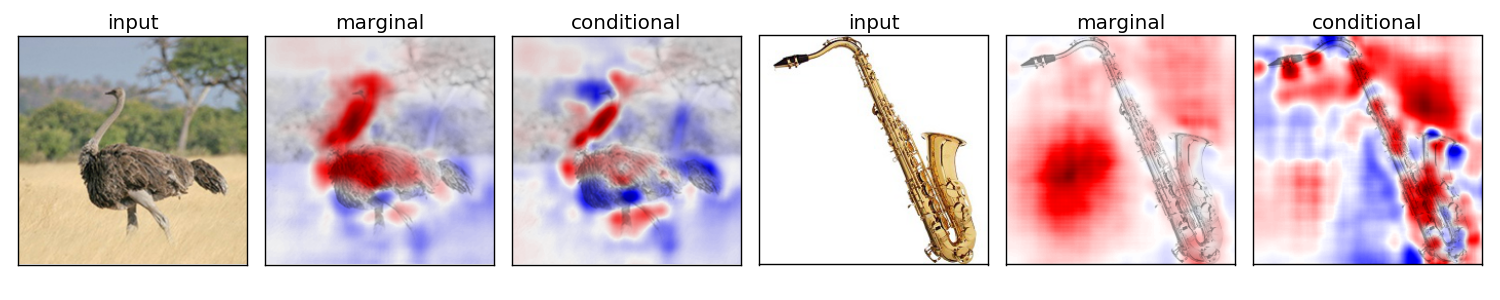}
\centering
\caption{\small \textbf{Visualization of the effects of marginal versus conditional sampling} using the GoogLeNet classifier. The classifier makes correct predictions (ostrich and saxophone), and we show the evidence for (red) and against (blue) this decision at the output layer. We can see that conditional sampling gives more targeted explanations compared to marginal sampling. Also, marginal sampling assigns too much importance on pixels that are easily predictable conditioned on their neighboring pixels.}
\label{fig:experiments:marg-vs-cond}
\end{figure}

\vspace{0.2cm}
\textit{Marginal vs Conditional Sampling}

Figure \ref{fig:experiments:marg-vs-cond} shows visualizations of the spatial support for the highest scoring class, using marginal and conditional sampling (with $k=10$ and $l=14$). We can see that conditional sampling leads to results that are more refined in the sense that they concentrate more around the object. We can also see that marginal sampling leads to pixels being declared as important that are very easily predictable conditioned on their neighboring pixels (like in the saxophone example). Throughout our experiments, we have found that conditional sampling tends to give more specific and fine-grained results than marginal sampling. For the rest of our experiments, we therefore show results using conditional sampling only.

\vspace{0.2cm}
\textit{Multivariate Analysis}

\begin{figure}
\includegraphics[width=0.95\textwidth]{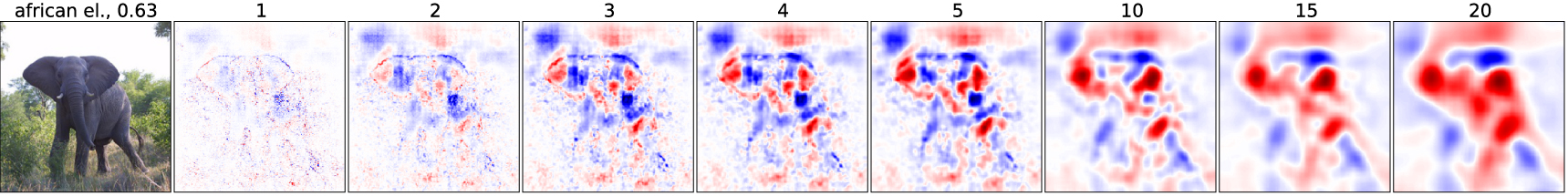}
\centering
\caption{\small \textbf{Visualization of how different window sizes influence the visualization result.} We used the conditional sampling method and the AlexNet classifier with $l=k+4$ and varying $k$. We can see that even when removing single pixels ($k=1$), this has a noticeable effect on the classifier and more important pixels get a higher score. By increasing the window size we can get a more easily interpretable, smooth result until the image gets blurry for very large window sizes.}
\label{fig:experiments:imagenet-winsize}
\end{figure}

For ImageNet data, we have observed that setting $k=10$ gives a good trade-off between sharp results and a smooth appearance. Figure \ref{fig:experiments:imagenet-winsize} shows how different window sizes influence the resolution of the visualization. Surprisingly, removing only one pixel does have a measurable effect on the prediction, and the largest effect comes from sensitive pixels. We expected that removing only one pixel does not have any effect on the classification outcome, but apparently the classifier is sensitive even to these small changes. However when using such a small window size, it is difficult to make sense of the sign information in the visualization. If we want to get a good impression of which parts in the image are evidence for/against a class, it is therefore better to use larger windows. If $k$ is chosen too large however, the results tend to get blurry. Note that these results are not just simple averages of one another, but a multivariate approach is indeed necessary to observe the presented results. 

\vspace{0.2cm}
\textit{Deep Visualization of Hidden Network Layers}

\begin{figure*}
\includegraphics[width=0.99\textwidth]{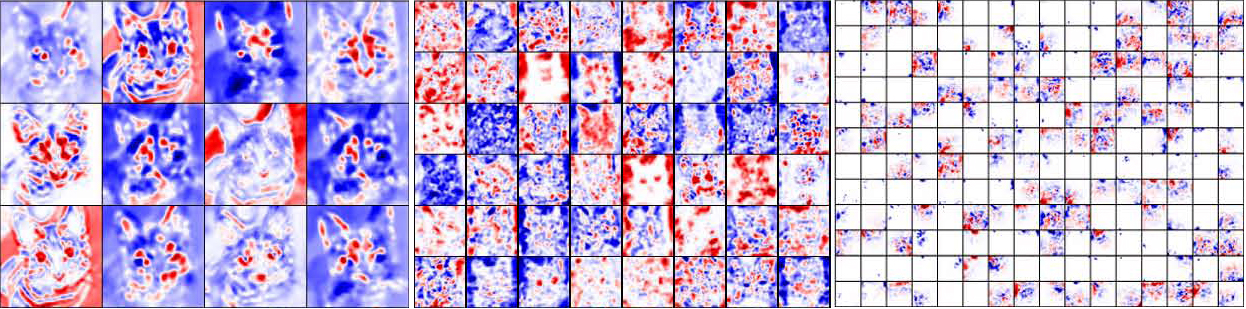}
\centering
\caption{\small Visualization of feature maps from thee different layers of the GoogLeNet (l.t.r.: ''conv1/7x7\_s2'', ''inception\_3a/output'', ''inception\_5b/output''), using conditional sampling and patch sizes $k=10$ and $l=14$ (see alg. \ref{alg:pred-diff}). For each feature map in the convolutional layer, we first evaluate the relevance for every single unit, and then average the results over all the units in one feature map to get a sense of what the unit is doing as a whole. \textit{Red} pixels activate a unit, \textit{blue} pixels decreased the activation.}
\label{fig:experiments:deepvis-1}
\end{figure*}

\begin{figure}
\includegraphics[width=0.7\textwidth]{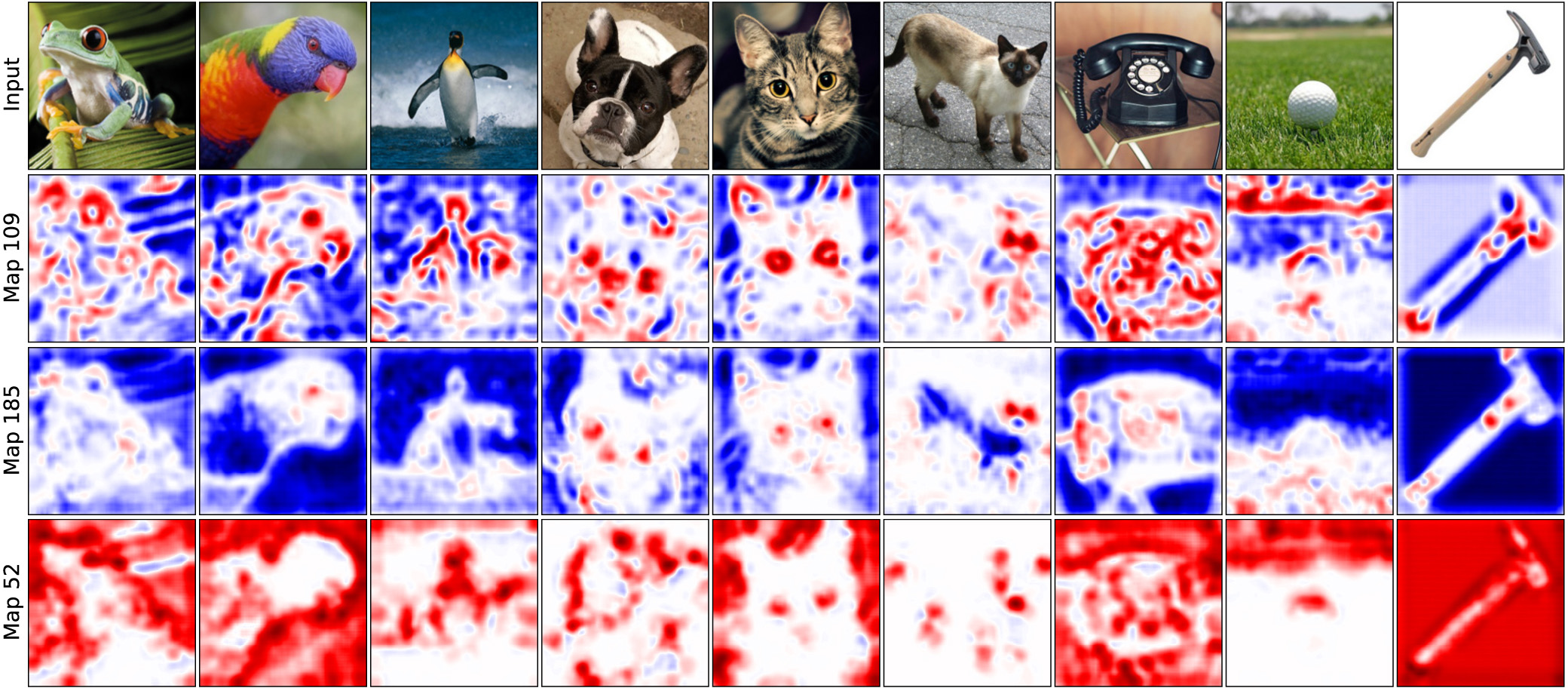}
\centering
\caption{\small \textbf{Visualization of three different feature maps}, taken from the ''inception\_3a/output'' layer of the GoogLeNet (from the middle of the network). Shown is the average relevance of the input features over all activations of the feature map. We used patch sizes $k=10$ and $l=14$ (see alg. \ref{alg:pred-diff}). \textit{Red} pixels activate a unit, \textit{blue} pixels decreased the activation.}
\label{fig:experiments:deepvis-2}
\end{figure}

Our third main contribution is the extension of the method to neural networks; to understand the role of hidden layers in a DNN. Figure \ref{fig:experiments:deepvis-1} shows how different feature maps in three different layers of the GoogLeNet react to the input of a tabby cat (see figure \ref{fig:experiments:deepvis-2}, middle image). For each feature map in a convolutional layer, we first compute the relevance of the input image for each hidden unit in that map. To estimate what the feature map as a whole is doing, we show the average of the relevance vectors over all units in that feature map.
The first convolutional layer works with different types of simple image filters (e.g., edge detectors), and what we see is which parts of the input image respond positively or negatively to these filters. The layer we picked from somewhere in the middle of the network is specialized to higher level features (like facial features of the cat).
The activations of the last convolutional layer are very sparse across feature channels, indicating that these units are highly specialized. 

To get a sense of what single feature maps in convolutional layers are doing, we can look at their visualization for different input images and look for patterns in their behavior. 
Figure \ref{fig:experiments:deepvis-2} shows this for four different feature maps from a layer from the middle of the GoogLeNet network. We can directly see which kind of features the model has learned at this stage in the network. For example, one feature map is mostly activated by the eyes of animals (third row), and another is looking mostly at the background (last row).

\vspace{0.2cm}
\textit{Penultimate vs Output Layer}

\begin{figure}
\includegraphics[width=0.99\textwidth]{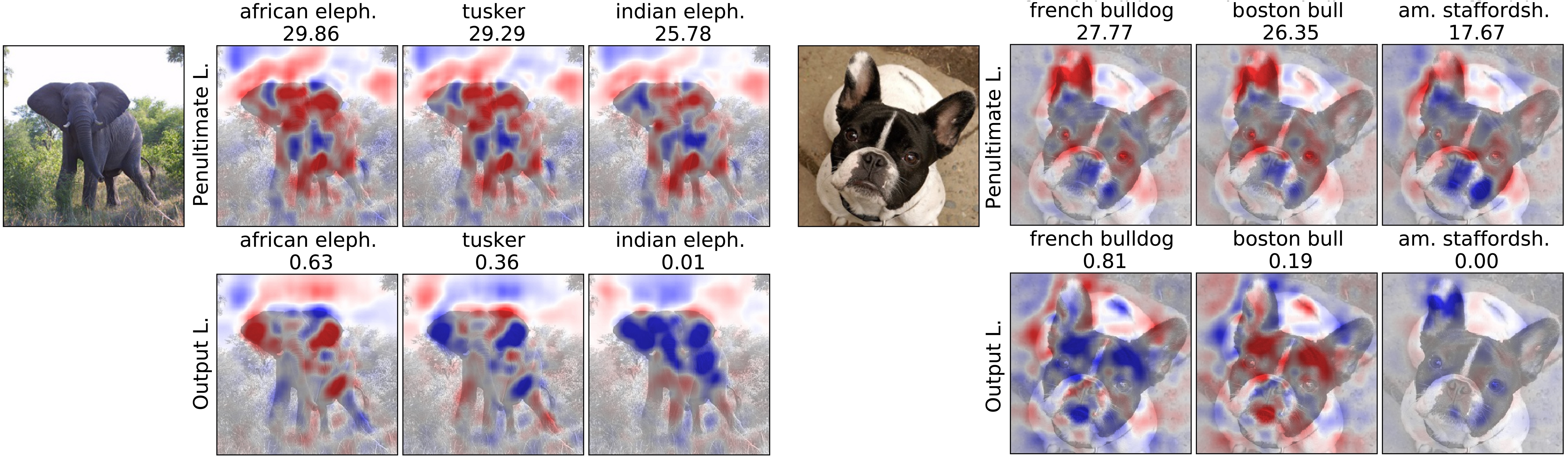} 
\centering
\caption{\small \textbf{Visualization of the support for the top-three scoring classes in the penultimate- and output layer}. Next to the input image, the first row shows the results with respect to the penultimate layer; the second row with respect to the output layer. For each image, we additionally report the values of the units. We used the AlexNet with conditional sampling and patch sizes $k=10$ and $l=14$ (see alg. \ref{alg:pred-diff}). \textit{Red} pixels are evidence for a class, and \textit{blue} against it.}
\label{fig:experiments:pen-vs-out}
\end{figure}

If we visualize the influence of the input features on the penultimate (pre-softmax) layer, we show only the evidence for/against this particular class, without taking other classes into consideration. After the softmax operation however, the values of the nodes are all interdependent: a drop in the probability for one class could be due to less evidence for it, or because a different class becomes more likely.
Figure \ref{fig:experiments:pen-vs-out} compares visualizations for the last two layers. By looking at the top three scoring classes, we can see that the visualizations in the penultimate layer look very similar if the classes are similar (like different dog breeds).
When looking at the output layer however, they look rather different. Consider the case of the elephants: the top three classes are different elephant subspecies, and the visualizations of the penultimate layer look similar since every subspecies can be identified by similar characteristics. But in the output layer, we can see how the classifier decides for one of the three types of elephants and against the others: the ears in this case are the crucial difference. 

\vspace{0.2cm}
\textit{Network Comparison}

\begin{figure*}
\includegraphics[width=0.99\textwidth]{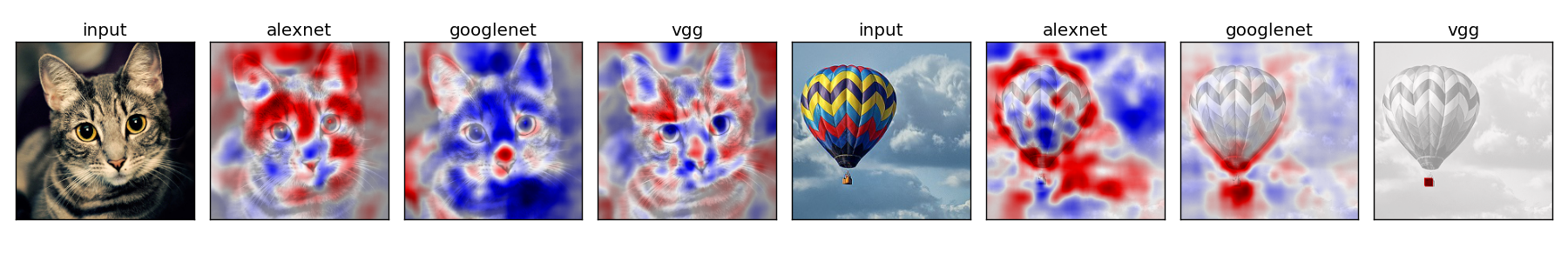}
\centering
\caption{\small \textbf{Comparison of the prediction visualization of different DCNN architectures.} For two input images, we show the results of the prediction difference analysis when using different neural networks - the AlexNet, GoogLeNet and VGG network.}
\label{fig:experiments:imagenet-network-comparison}
\end{figure*}

When analyzing how neural networks make decisions, we can also compare how different network architectures influence the visualization. Here, we tested our method on the AlexNet, the GoogLeNet and the VGG network. Figure \ref{fig:experiments:imagenet-network-comparison} shows the results for the three different networks, on two input images. The AlexNet seems to more on contextual information (the sky in the balloon image), which could be attributed to it having the least complex architecture compared to the other two networks. It is also interesting to see that the VGG network deems the basket of the balloon as very important compared to all other pixels. The second highest scoring class in this case was a parachute - presumably, the network learned to not confuse a balloon with a parachute by detecting a square basket (and not a human).


\subsection{MRI data: Explaining Classifier Decisions in Medical Imaging} \label{sec:experiments:mri}

To illustrate how our visualization method can also be useful in a medical domain, we show some experimental results on an MRI dataset of HIV and healthy patients. In such settings, it is crucial that the practitioner has some insight into the algorithm's decision when classifying a patient, to weigh this information and incorporate it in the overall diagnosis process.

The dataset used here is referred to as the COBRA dataset. It contains 3D MRIs from 100 HIV patients and 70 healthy individuals, included in the Academic Medical Center (AMC) in Amsterdam, The Netherlands. 
Of these subjects, diffusion weighted MRI data were acquired. Preprocessing of the data was performed with software developed in-house, using the HPCN-UvA Neuroscience Gateway and using resources of the Dutch e-Science Grid \cite{Shahand2014}.
As a result, Fractional Anisotropy (FA) maps were computed.
FA is sensitive to microstructural damage and therefore expected to be, on average, decreased in patients.
Subjects were scanned on two 3.0 Tesla scanner systems, 121 subjects on a Philips Intera system and 39 on a Philips Ingenia system.
Patients and controls were evenly distributed.
FA images were spatially normalized to standard space \cite{Andersson2007a}, resulting in volumes with $91 \times 109 \times 91=902,629$ voxels.

We trained an L2-regularized Logistic Regression classifier on a subset of the MRI slices (slices 29-40 along the first axis) and on a balanced version of the dataset (by taking the first $70$ samples of the HIV class) to achieve an accuracy of $69.3$\% in a $10$-fold cross-validation test. Analyzing one image took around half an hour (on a CPU, with $k=3$ and $l=7$, see algorithm \ref{alg:pred-diff}). For conditional sampling, we also tried adding location information in equation (\ref{eq:marginal-approx}), i.e., we split up the 3D image into a $20\times20\times20$ grid and also condition on the index in that grid. We found that this slightly improved the interpretability of the results, since the pixel values in the special case of MRI scans does depend on spacial location as well.

\begin{figure}
\includegraphics[width=0.65\textwidth]{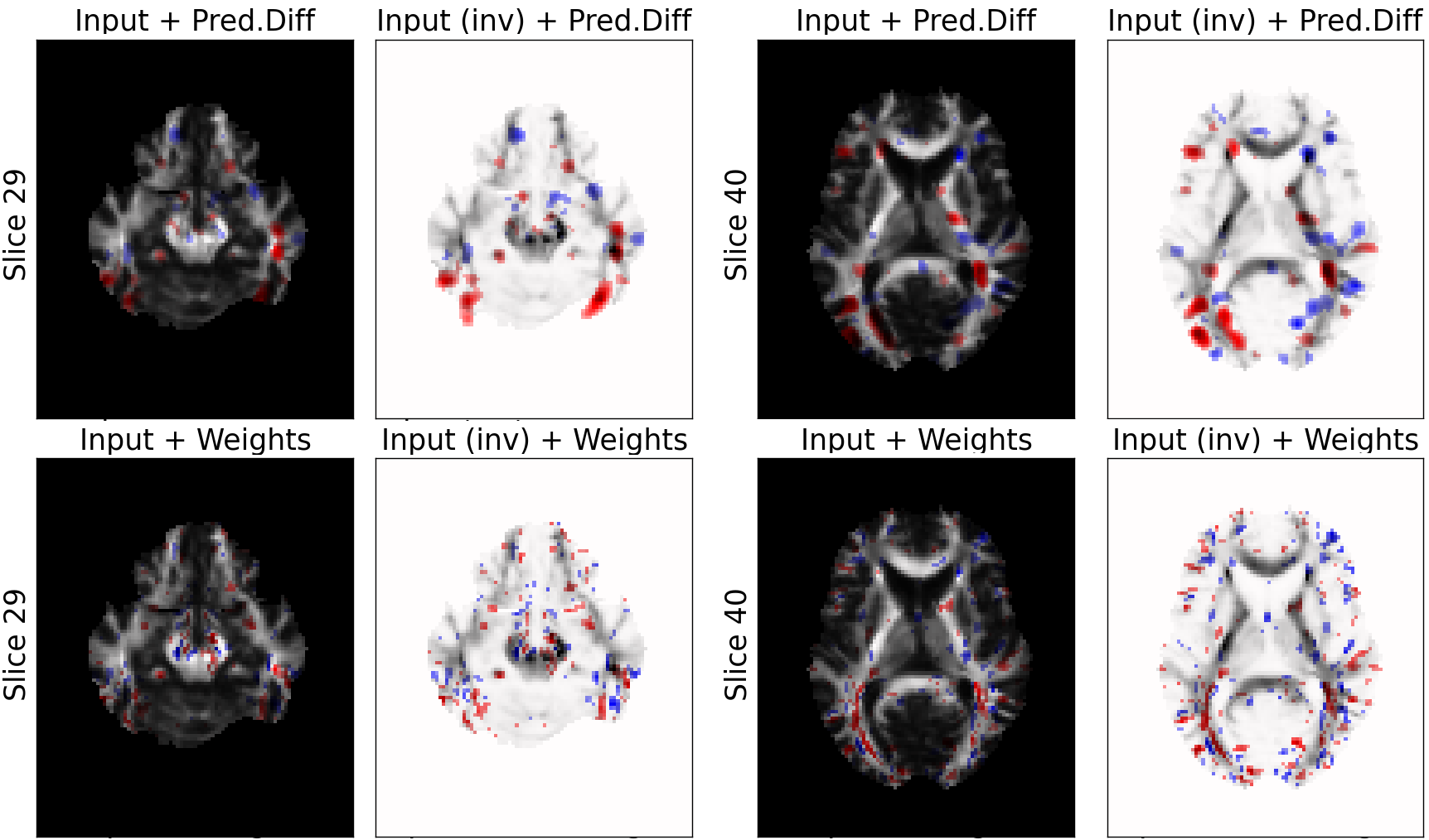}
\centering
\caption{\small \textbf{Visualization of the support for the correct classification ''HIV'', using the Prediction Difference method and Logistic Regression Weights.} For an HIV sample, we show the results  with the prediction difference (first row), and using the weights of the logistic regression classifier (second row), for slices $29$ and $40$ (along the first axis). \textit{Red} are positive values, and \textit{blue} negative. For each slice, the left image shows the original image, overlaid with the relevance values. The right image shows the original image with reversed colors and the relevance values. Relevance values are shown only for voxels with (absolute) relevance value above $15$\% of the (absolute) maximum value.} 
\label{fig:experiments:mri-color}
\end{figure}

\begin{figure}
\includegraphics[width=0.9\textwidth]{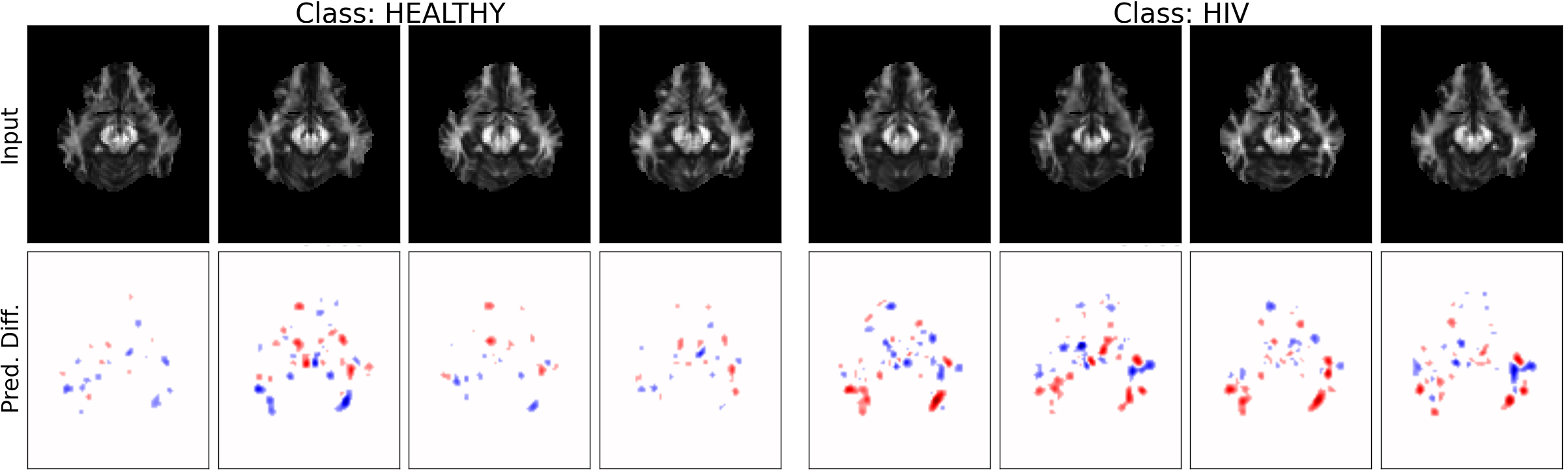}
\centering
\caption{\small \textbf{Prediction difference visualization for different samples.} The first four samples are of the class ''healthy''; the last four of the class ''HIV''. All images show slice 39 (along the first axis). All samples are correctly classified, and the results show evidence for (red) and against (blue) this decision. Prediction differences are shown only for voxels with (absolute) relevance value above $15$\% of the (absolute) maximum value.}
\label{fig:experiments:mri-classes}
\end{figure}

\begin{figure}
\includegraphics[width=0.8\textwidth]{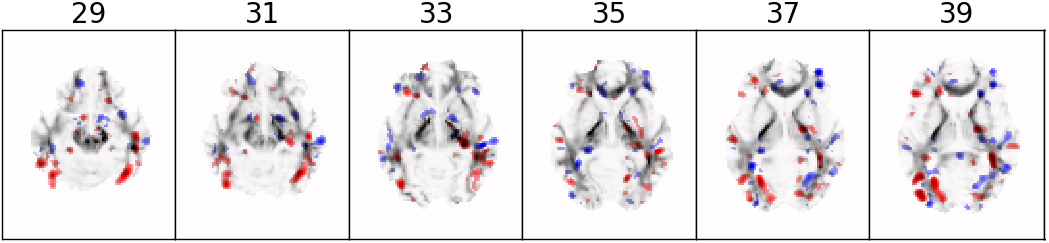} 
\centering
\caption{\small \textbf{Visualization results across different slices of the MRI image}, using the same input image as shown in \ref{fig:experiments:mri-color}. Prediction differences are shown only for voxels with (absolute) relevance value above $15$\% of the (absolute) maximum value.}
\label{fig:experiments:mri-slices}
\end{figure}

\begin{figure}
\includegraphics[width=0.8\textwidth]{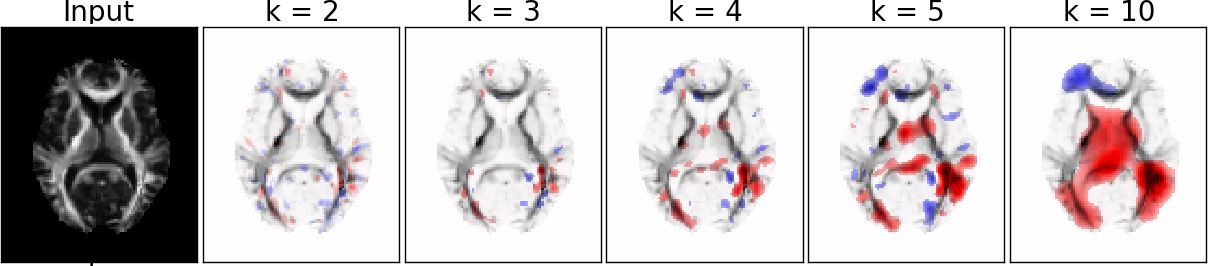}
\centering
\caption{\small \textbf{How the patch size influences the visualization.} For the input image (HIV sample, slice $39$ along the first axis) we show the visualization with different patch sizes ($k$ in alg. \ref{alg:pred-diff}). Prediction differences are shown only for voxels with (absolute) relevance value above $15$\% of the (absolute) maximum (for $k=2$ it is $10$\%).}
\label{fig:experiments:mri-winsize}
\end{figure}

Figure \ref{fig:experiments:mri-color} (first row) shows one way via which the prediction difference results could be presented to a physician, for an HIV sample. By overlapping the prediction difference and the MRI image, the exact regions can be pointed out that are evidence for (red parts) or against (blue parts) the classifier's decision. The second row shows the results using the weights of the logistic regression classifier, which is a commonly used method in neuroscientific literature. We can see that they are considerably noisier (in the sense that, compared to our method, the voxels relevant for the classification decisions are more scattered), and also, they are not specific to the given image.
Figure \ref{fig:experiments:mri-classes} shows the visualization results for four healthy, and four HIV samples. We can clearly see that the patterns for the two classes are distinct, and there is some pattern to the decision of the classifier, but which is still specific to the input image.
Figure \ref{fig:experiments:mri-slices} shows the same (HIV) sample as in figure \ref{fig:experiments:mri-color} along different axes, and
figure \ref{fig:experiments:mri-winsize} shows how the visualization changes with different patch sizes. We believe that both varying the slice and patch size can give different insights to a clinician, and in clinical practice, a $3$D animation where these parameters can be adjusted would be very useful for analyzing the visualization result.

In general we can assume that the better the classifier, the closer the explanations for its decisions are to the true class difference. For clinical practice it is therefore crucial to have very good classifiers. This will increase computation time, but in many medical settings, longer waiting times for test results are common and worth the wait if the patient is not in an acute life threatening condition (e.g., when predicting HIV or Alzheimer from MRI scans, or the field of cancer diagnosis and detection). The presented results here are for demonstration purposes of the visualization method, and we claim no medical validity. A thorough qualitative analysis incorporating expert knowledge was outside the scope of this paper.


\section{Future work} 

In our experiments, we used a simple multivariate normal distribution for conditional sampling. We can imagine that using more sophisticated generative models will lead to better results: pixels that are easily predictable by their surrounding are downweighted even more. However this will also significantly increase the computational resources needed to produce the explanations. Similarly, we could try to modify equation (\ref{eq:conditional-approx}) to get an even better approximation by using a conditional distribution that takes more information about the whole image into account (like adding spatial information for the MRI scans).

To make the method applicable for clinical analysis and practice, a better classification algorithm is required. Also, software that visualizes the results as an interactive $3$D model will improve the usability of the system.

\section{Conclusion}

We presented a new method for visualizing deep neural networks that improves on previous methods by using a more powerful conditional, multivariate model.
The visualization method shows which pixels of a specific input image are evidence for or against a node in the network. 
The signed information offers new insights - for research on the networks, as well as the acceptance and usability in domains like healthcare. 
While our method requires significant computational resources, real-time 3D visualization is possible when visualizations are pre-computed.
With further optimization and powerful GPUs, pre-computation time can be reduced a lot further.
In our experiments, we have presented several ways in which the visualization method can be put into use for analyzing how DCNNs make decisions.

\newpage
\small
\subsubsection*{Acknowledgments} 

This work was supported by AWS in Education Grant award. We thank Facebook and Google for financial support, and our reviewers for their time and valuable, constructive feedback.

This work was also in part supported by: Innoviris, the Brussels Institute for Research and Innovation, Brussels, Belgium; the Nuts-OHRA Foundation (grant no. 1003-026), Amsterdam, The Netherlands; The Netherlands Organization for Health Research and Development (ZonMW) together with AIDS Fonds (grant no 300020007 and 2009063). Additional unrestricted scientific grants were received from Gilead Sciences, ViiV Healthcare, Janssen Pharmaceutica N.V., Bristol-Myers Squibb, Boehringer Ingelheim, and Merck\&Co. 

{\small
We thank Barbara Elsenga, Jane Berkel, Sandra Moll, Maja Totté, and Marjolein Martens for running the AGEhIV study program and capturing our data with such care and passion. We thank Yolanda Ruijs-Tiggelman, Lia Veenenberg-Benschop, Sima Zaheri, and Mariska Hillebregt at the HIV Monitoring Foundation for their contributions to data management. We thank Aafien Henderiks and Hans-Erik Nobel for their advice on logistics and organization at the Academic Medical Center. We thank all HIV-physicians and HIV-nurses at the Academic Medical Center for their efforts to include the HIV-infected participants into the AGEhIV Cohort Study, and the Municipal Health Service Amsterdam personnel for their efforts to include the HIV-uninfected participants into the AGEhIV Cohort Study. We thank all study participants without whom this research would not be possible.

\textbf{AGEhIV Cohort Study Group.}
Scientific oversight and coordination: P. Reiss (principal investigator), F.W.N.M. Wit, M. van der Valk, J. Schouten, K.W. Kooij, R.A. van Zoest, E. Verheij, B.C. Elsenga (Academic Medical Center (AMC), Department of Global Health and Amsterdam Institute for Global Health and Development (AIGHD)). 
M. Prins (co-principal investigator), M.F. Schim van der Loeff, M. Martens, S. Moll, J. Berkel, M. Totté, G.R. Visser, L. May, S. Kovalev, A. Newsum, M. Dijkstra (Public Health Service of Amsterdam, Department of Infectious Diseases). 
Datamanagement: S. Zaheri, M.M.J. Hillebregt, Y.M.C. Ruijs, D.P. Benschop, A. el Berkaoui (HIV Monitoring Foundation).
Central laboratory support: N.A. Kootstra, A.M. Harskamp-Holwerda, I. Maurer, T. Booiman, M.M. Mangas Ruiz, A.F. Girigorie, B. Boeser-Nunnink (AMC, Laboratory for Viral Immune Pathogenesis and Department of Experimental Immunology).
Project management and administrative support: W. Zikkenheiner, F.R. Janssen (AIGHD).
Participating HIV physicians and nurses: S.E. Geerlings, M.H. Godfried, A. Goorhuis, J.W.R. Hovius, J.T.M. van der Meer, F.J.B. Nellen, T. van der Poll, J.M. Prins, P. Reiss, M. van der Valk, W.J. Wiersinga, M. van Vugt, G. de Bree, F.W.N.M. Wit; J. van Eden, A.M.H. van Hes, M. Mutschelknauss , H.E. Nobel, F.J.J. Pijnappel, M. Bijsterveld, A. Weijsenfeld, S. Smalhout (AMC, Division of Infectious Diseases).
Other collaborators: J. de Jong, P.G. Postema (AMC, Department of Cardiology); P.H.L.T. Bisschop, M.J.M. Serlie (AMC, Division of Endocrinology and Metabolism);  P. Lips (Free University Medical Center Amsterdam); E. Dekker (AMC, Department of Gastroenterology); N. van der Velde (AMC, Division of Geriatric Medicine); J.M.R. Willemsen, L. Vogt (AMC, Division of Nephrology); J. Schouten, P. Portegies, B.A. Schmand, G.J. Geurtsen (AMC, Department of Neurology);  F.D. Verbraak, N. Demirkaya (AMC, Department of Ophthalmology); I. Visser (AMC, Department of Psychiatry); A. Schadé (Free University Medical Center Amsterdam, Department of Psychiatry); P.T. Nieuwkerk, N. Langebeek (AMC, Department of Medical Psychology); R.P. van Steenwijk, E. Dijkers (AMC, Department of Pulmonary medicine); C.B.L.M. Majoie, M.W.A. Caan, T. Su (AMC, Department of Radiology); H.W. van Lunsen, M.A.F. Nievaard (AMC, Department of Gynaecology); B.J.H. van den Born, E.S.G. Stroes, (AMC, Division of Vascular Medicine); W.M.C. Mulder (HIV Vereniging Nederland).
}

\bibliography{iclr2017_conference}
\bibliographystyle{iclr2017_conference}

\newpage
\appendix

\section{Random Results} \label{app:random_results}

\begin{figure}[h!]
\includegraphics[width=\textwidth]{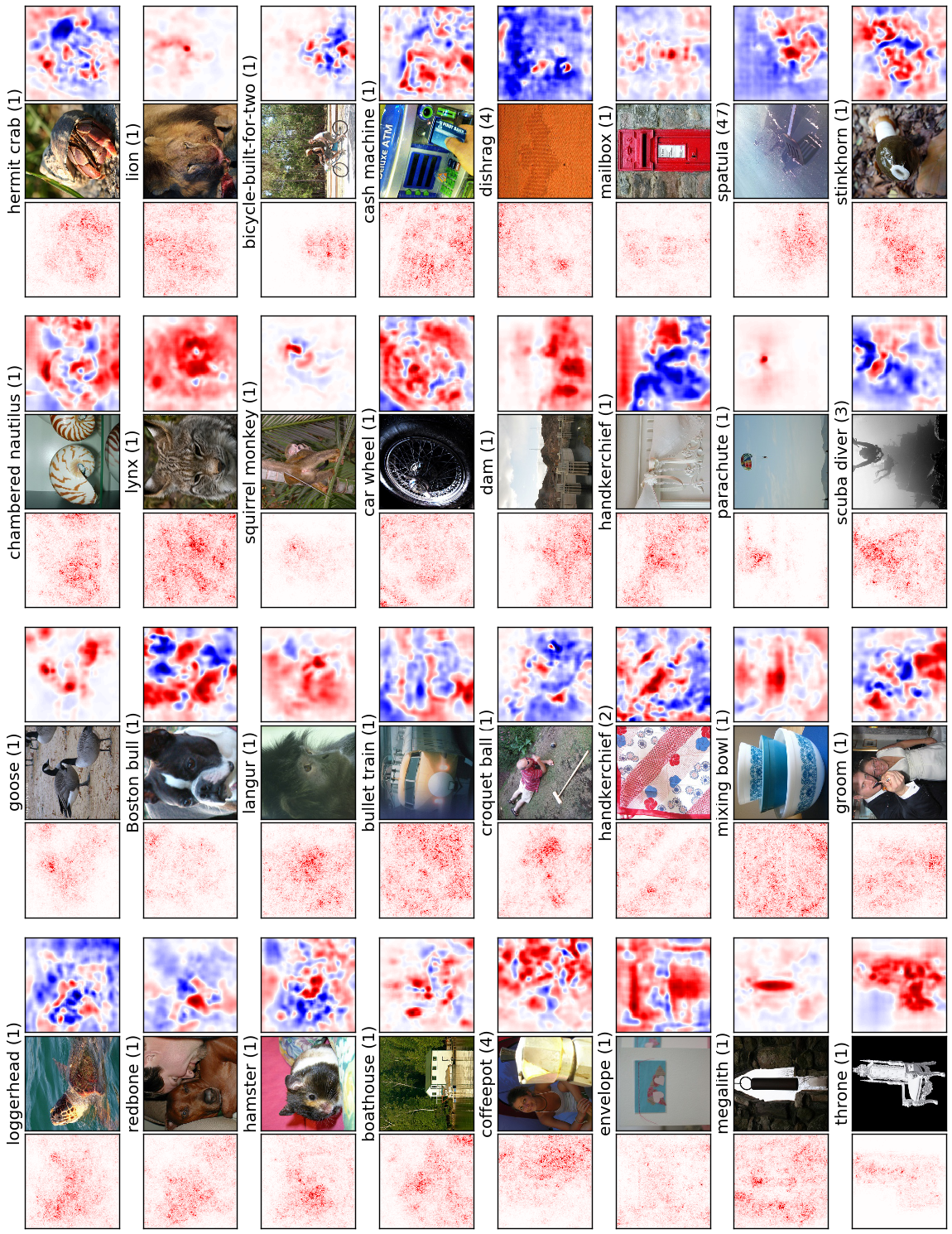}
\centering
\caption{\small \textbf{Results on 34 randomly chosen ImageNet images}. Middle columns: original image; left columns: sensitivity maps \citep{simonyan2013deep} where the red pixels indicate high sensitivity, and white pixels mean no sensitivity (note that we show the absolute values of the partial derivatives, since the sign cannot be interpreted like in our method); right columns: results from our method. For both methods, we visualize the results with respect to the correct class which is given above the image. In brackets we see how the classifier ranks this class, i.e., a ($1$) means it was correctly classified, whereas a ($4$) means that it was misclassified, and the correct class was ranked fourth. For our method, red areas show evidence for the correct class, and blue areas show evidence against the class (e.g., the scuba diver looks more like a tea pot to the classifier).
}
\label{fig:app:random}
\end{figure}

\end{document}